\documentclass{article}

\usepackage[pdftex]{graphicx}
\usepackage{amsmath}
\usepackage{amssymb}
\usepackage[pagebackref=true,breaklinks=true,colorlinks,bookmarks=false]{hyperref}

\title{How to represent part-whole hierarchies\\
 in a neural network}
\author{Geoffrey Hinton\\
Google Research\\
\&\\
The Vector Institute\\
\&\\
Department of Computer Science\\
University of Toronto}
\date{February 22, 2021}

\begin{document}

\maketitle
\begin{abstract}
This paper does not describe a working system. Instead, it presents a single idea about representation which allows advances made by several different groups to be combined into an imaginary system called GLOM\footnote{GLOM is derived from the slang "glom together" which may derive from the word "agglomerate".}. The advances include transformers, neural fields,  contrastive representation learning, distillation and capsules. GLOM answers the question: How can a neural network with a fixed architecture parse an image into a part-whole hierarchy which has a different structure for each image? The idea is simply to use islands of identical vectors to represent the nodes in the parse tree. If GLOM can be made to work, it should significantly improve the interpretability of the representations produced by transformer-like systems when applied to vision or language.
\end{abstract}
 \section{Overview of the idea}
There is strong psychological evidence that people parse visual scenes into part-whole hierarchies and model the viewpoint-invariant spatial relationship between a part and a whole as the coordinate transformation between intrinsic coordinate frames that they assign to the part and the whole~\cite{HintonCube}.  If we want to make neural networks that understand images in the same way as people do, we need to figure out how neural networks can represent part-whole hierarchies. This is difficult because a real neural network cannot dynamically allocate a group of neurons to represent a node in a parse tree\footnote{What neurons do is determined by their incoming and outgoing weights and real neurons cannot completely change these weights rapidly.}. The inability of neural nets to dynamically allocate neurons was the motivation for a series of models that used ``capsules''~\cite{capsules2017,capsules2018,capsules2019}. These models made the assumption that a group of neurons called a capsule would be permanently dedicated to a part of a particular type occurring in a particular region of the image. A parse tree could then be created by activating a subset of these pre-existing, type-specific capsules and the appropriate connections between them.  This paper describes a very different way of using capsules to represent the part-whole hierarchy in a neural net.

Even though this paper is primarily concerned with the perception of a single static image, GLOM is most easily understood as a pipeline for processing a sequence of frames, so a static image will be treated as a sequence of identical frames.

\begin{figure}[h!]
\centerline{\includegraphics[width=5in]{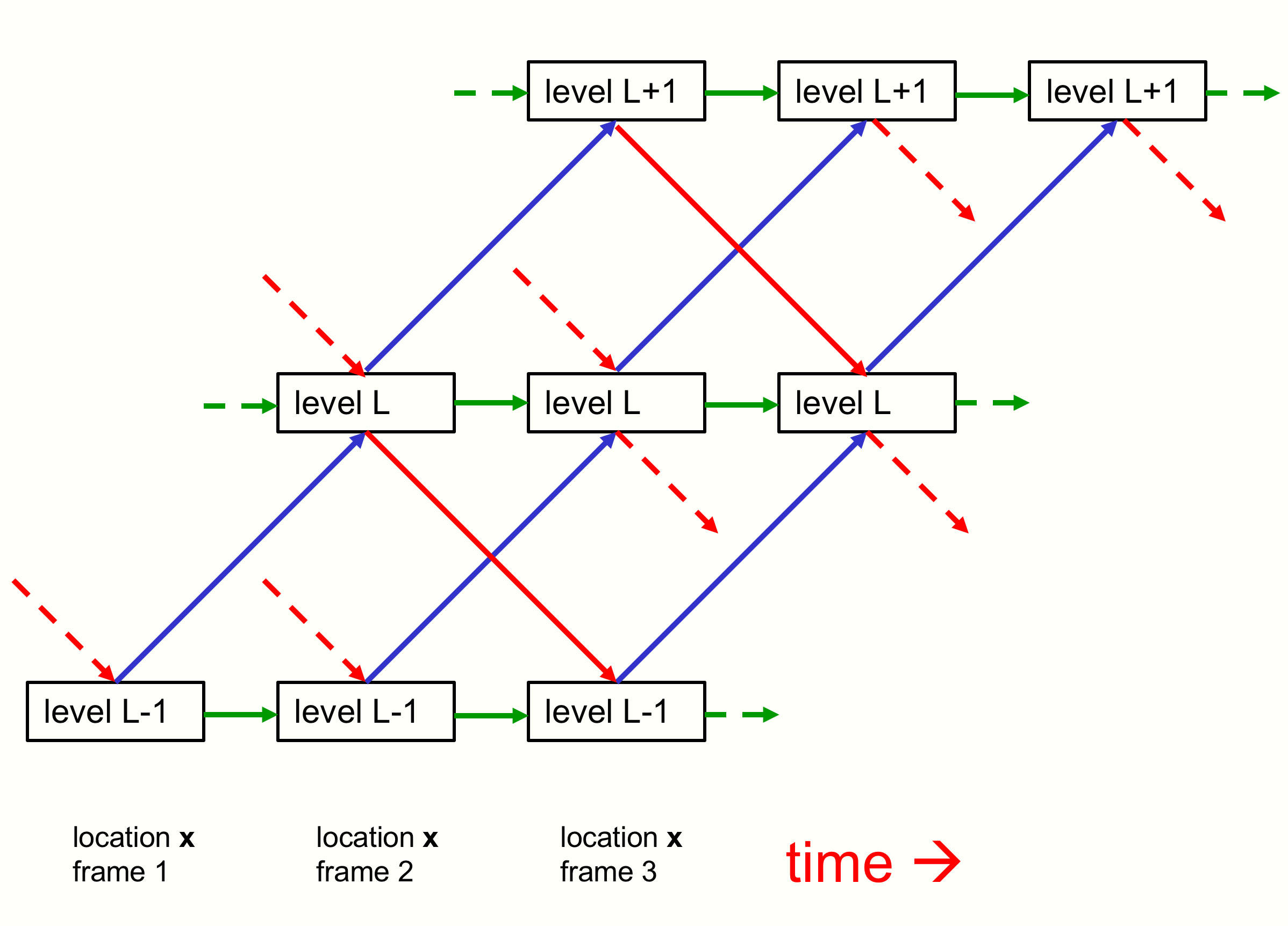}}
\caption{Showing the bottom-up, top-down, and same level interactions among three adjacent levels of the proposed GLOM architecture {\it for a single column}. The blue and red arrows representing bottom-up and top-down interactions are implemented by two different neural networks that have several hidden layers. These networks can differ between pairs of levels but they are shared across columns and across time-steps.  The top-down net should probably use sinusoidal units\cite{sitzmann2020implicit}. For a static image, the green arrows could simply be scaled residual connections that implement temporal smoothing of the embedding at each level. For video, the green connections could be neural networks that learn temporal dynamics based on several previous states of the capsule.  Interactions between the embedding vectors at the same level in different columns are implemented by a non-adaptive, attention-weighted, local smoother which is not shown. 
}
\label{fig:newglomarchitecture}
\end{figure}

The GLOM architecture\footnote{The GLOM architecture has some similarity to models that use the errors in top-down predictions as their bottom-up signals~\cite{RaoBallard}, but in a nonlinear system the bottom-up signals cannot just carry the prediction error because the full activity vector is required to select the right operating regime for the non-linear units.} is composed of a large number of columns\footnote{Each level in a column bears some resemblance to a hypercolumn as described by neuroscientists.} which all use exactly the same weights. Each column is a stack of spatially local autoencoders that learn multiple levels of representation for what is happening in a small image patch. Each autoencoder transforms the embedding at one level into the embedding at an adjacent level using a multilayer bottom-up encoder and a multilayer top-down decoder. These levels correspond to the levels in a part-whole hierarchy. When shown an image of a face, for example, a single column might converge on embedding vectors\footnote{An embedding vector is the activity vector of a capsule.} representing a nostril, a nose, a face, and a person. Figure \ref{fig:newglomarchitecture} shows how the embeddings at different levels interact in a single column. 

Figure \ref{fig:newglomarchitecture} does not show the interactions between embeddings at the same level in different columns. These are much simpler than the interactions within a column because they do not need to implement part-whole coordinate transforms. They are like the attention-weighted interactions between columns representing different word fragments in a multi-headed transformer~\cite{BERT}, but they are simpler because the query, key and value vectors are all identical to the embedding vector. The role of the inter-column interactions is to produce islands of identical embeddings at a level by making each embedding vector at that level regress towards other similar vectors at nearby locations.  This creates multiple local "echo chambers" in which embeddings at a level attend mainly to other like-minded embeddings.  

At each discrete time and in each column separately, the embedding at a level is updated to be the weighted average of four contributions:
\begin{enumerate}
    \item The prediction produced by the bottom-up neural net acting on the embedding at the level below at the previous time. 
    \item The prediction produced by the top-down neural net acting on the embedding at the level above at the previous time.
    \item The embedding vector at the previous time step. 
    \item The attention-weighted average of the embeddings at the same level in nearby columns at the previous time.
\end{enumerate}

\begin{figure}[h!]
\vspace*{1.5in}
\centerline{\includegraphics[width=5in]{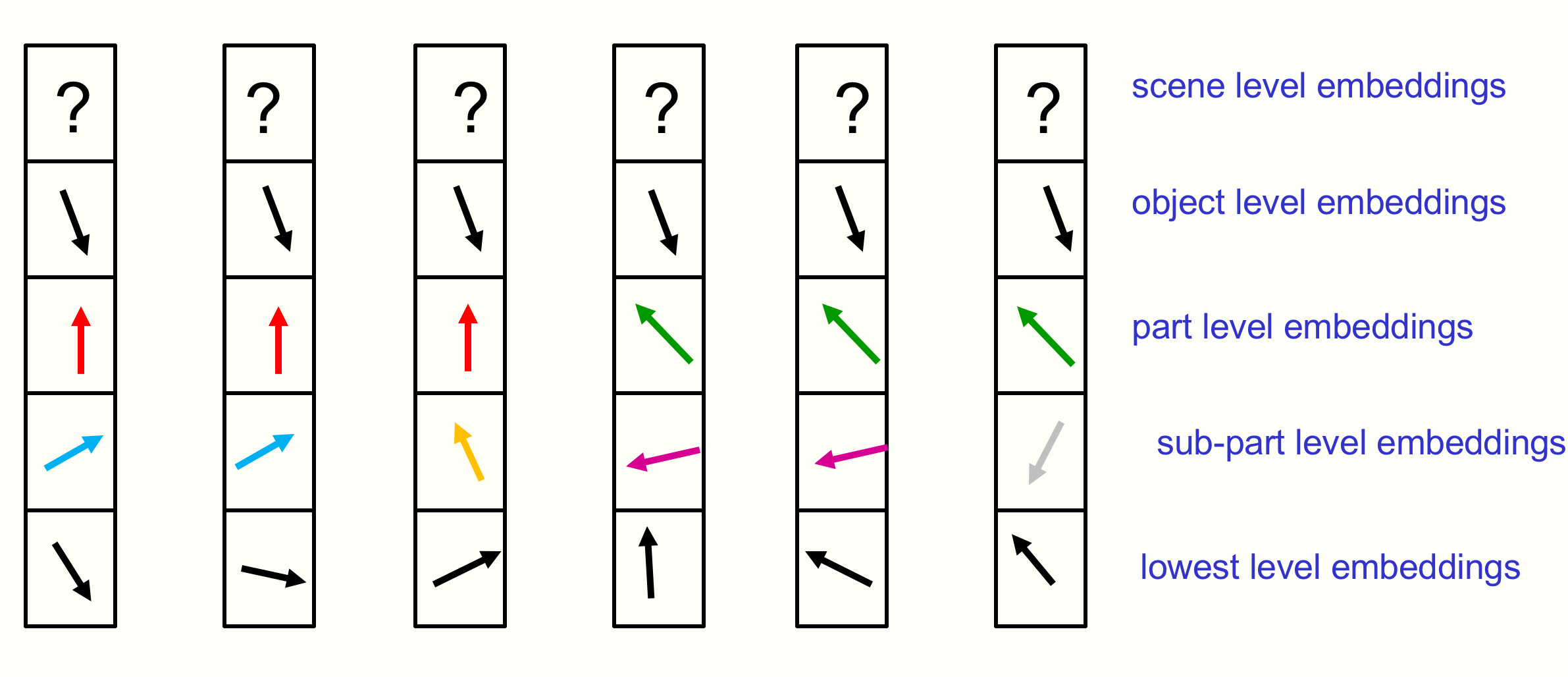}}
\caption{A picture of the embeddings {\it at a particular time} in six nearby columns. All of the locations shown belong to the same object and the scene level has not yet settled on a shared vector. The complete embedding vector for each location is shown by dividing the vector into a separate section for each level in the part-whole hierarchy and then showing the high-dimensional embedding vector for a level as a 2-D vector. This makes it easy to illustrate alignment of the embedding vectors of different locations. The islands of identical vectors at the various levels shown in the figure represent a parse tree. But islands of identity are considerably more powerful than phrase structure grammars. They have no difficulty representing disconnected objects as in "Will this slow phrase structure grammarians down?"}
\label{fig:islands}
\end{figure}

For a static image, the embeddings at a level should settle down over time to produce distinct islands of nearly identical vectors. These islands should be larger at higher levels as shown in figure \ref{fig:islands}. Using the islands of similarity to represent the parse of an image avoids the need to allocate groups of neurons to represent nodes of the parse tree on the fly, or to set aside groups of neurons for all possible nodes in advance. Instead of allocating neural hardware to represent a node in a parse tree and giving the node pointers to its ancestor and descendants, GLOM allocates an appropriate activity vector to represent the node and uses the same activity vector for all the locations belonging to the node\footnote{The idea of using similarity of vectors to do segmentation has been used in earlier work on directional unit Boltzmann machines~\cite{ZemelWilliamsMozer}}.  The ability to access the ancestor and descendants of the node is implemented by the bottom-up and top down neural nets rather than by using RAM to do table look-up.

Like BERT~\cite{BERT}, the whole system can be trained end-to-end to reconstruct images at the final time-step from input images which have missing regions, but the objective function also includes two regularizers that encourage islands of near identical vectors at each level. The regularizers are simply the agreement between the new embedding at a level and the bottom-up and top-down predictions. Increasing this agreement facilitates the formation of local islands. 
\newpage

 \newpage

 \newpage
 \section{Introduction}
This paper proposes the idea of using islands of similar vectors to represent the parse tree of an image and then explores some of the many ramifications of this idea by describing an imaginary system called GLOM that implements it. It concludes with some speculations about how the brain might implement some aspects of GLOM. But first some disclaimers:

\noindent {\bf Disclaimer1:} Human vision is a sampling process in which intelligently chosen fixation points are used to acquire the information required to perform a task using retinas that have much higher resolution around the fixation point. The same neural circuitry is reused for each new fixation. For the purposes of this paper, I assume a single retina or camera with uniform resolution and only consider what happens on the first fixation.

\noindent {\bf Disclaimer 2:} To avoid cumbersome terms like ``sub-sub-parts'', I will often talk about parts and wholes as if there were only two levels in the part-whole hierarchy. But a section of the complete embedding vector that is called a whole when considering levels L-1 and L is also called a part when considering levels L and L+1. 

In a computer that has general purpose, random access memory, the obvious way to represent the part-whole hierarchy for a specific image is to create a graph structure for that particular image by dynamically allocating pieces of the memory to the nodes in the graph and giving each node pointers to the nodes it is connected to.  Combining this type of dynamically created graph with neural network learning techniques has recently shown great promise \cite{PSG}, but if the whole computer is a neural network, it is far less obvious how to represent part-whole hierarchies that are different for every image if we want the structure of the neural net to be identical for all images. If we allow three-way interactions in which the activity of one neuron gates the connection between two other neurons~\cite{HintonCanonical}, it is easy to make the connections dynamic, but it is still unclear how to dynamically create a graph structure without the ability to allocate neurons on the fly. It is especially difficult in a real neural net where the knowledge is in the connection weights, which cannot easily be copied.

One rather cumbersome solution to this problem is to set aside a group of neurons, called a capsule, for each {\it possible} type of object or part in each region of the image\footnote{These regions can be larger for higher level parts which are more diverse but occur more sparsely in any one image.} and to use a routing algorithm to dynamically connect a small subset of active capsules into a graph that represents the parse of the image at hand. The activities of neurons within a capsule can then represent properties of a part such as the pose or deformation of a particular mouth or face. 

With considerable effort, models that use capsules have achieved some successes in supervised and unsupervised learning on small datasets \cite{capsules2017,capsules2018,capsules2019}, but they have not scaled well to larger datasets \cite{BarhamIsard}. Capsules do not have the signature of really practical ideas like stochastic gradient descent or transformers which just {\it want} to work. The fundamental weakness of capsules is that they use a mixture to model the set of possible parts. This forces a hard decision about whether a car headlight and an eye are really different parts. If they are modeled by the same capsule, the capsule cannot predict the identity of the whole. If they are modeled by different capsules the similarity in their relationship to their whole cannot be captured.  

One way to avoid using a mixture for modeling the different types of part is to have a set of identical, ``universal'' capsules, each of which contains enough knowledge to model any type of part \cite{slots,RussNitish,CanonicalCaps}. This allows part identities to have distributed representations, which allows better sharing of knowledge between similar parts. In neuroscience terminology, identities are value-coded rather than place-coded. However, it creates a symmetry breaking problem in deciding which universal object-level capsule each part should be routed to\footnote{Adam Kosoriek suggested using universal capsules in 2019, but I was put off by the symmetry breaking issue and failed to realise the importance of this approach.}.  

A more radical version of universal capsules, which avoids both symmetry breaking and routing, is to pre-assign a universal capsule to every location in the image. These ubiquitous universal capsules can be used to represent whatever happens to be at that location. An even more profligate version is to dedicate several different levels of ubiquitous universal capsule to each location so that a location can belong to a scene, an object, a part and a sub-part simultaneously. This paper explores this profligate way of  representing the part-whole hierarchy. It was inspired by a biological analogy, a mathematical analogy, and recent work on neural scene representations \cite{ha2016generating,sitzmann2019NEURIPS}.

\subsection{The biological analogy}
All the cells in the body have a copy of the whole genome. It seems wasteful for brain cells to contain the instructions for behaving like liver cells but it is convenient because it gives every cell its own private access to whatever DNA it might choose to express.  Each cell has an expression intensity\footnote{I use the word ``intensity'' rather than the word ``level'' so as not to confuse scalar intensities with discrete levels in a part-whole hierarchy.} for each gene and the vector of expression intensities is similar for cells that form part of the same organ.  

The analogy with neural nets goes like this: Each location in the image corresponds to a biological cell. The complete embedding vector for a location is like the vector of gene expression intensities for a cell. The forward pass is like the developmental process that allows a new vector of gene expression intensities to be determined by the previous vectors of expression intensities.  Objects are like organs: They are a collection of locations whose embedding vectors are all very similar at a high level. Within an object, the embedding vectors may differ at lower levels that correspond to the parts of the object (see figure \ref{fig:islands}).

\subsection{The mathematical analogy}
The Kolmogorov-Arnold superposition theorem states that every multivariate continuous function can be represented as a superposition of continuous functions of one variable\footnote{This solves a version of Hilbert's 13th problem.}.  For example, multiplication can be represented as the sum of the logs of the individual arguments followed by exponentiation.  In machine learning terminology, when it comes to multi-argument functions, addition is all you need. This assumes, of course, that you can find the right single-argument functions to encode the arguments of the multivariate function you want to represent and then find the right function to
decode the sum.  Kolmogorov proved this can always be done but the encoder functions used for the proof are so bizarre that they are of no practical relevance to neural networks.

The theorem does, however, suggest an interesting approach to combining information coming from many different locations. Perhaps we can learn how to encode the information at each location in such a way that simply averaging the encodings from different locations is the only form of interaction we need\footnote{This has a resemblance to variational learning \cite{radford}, where we start by assuming that the log posterior distribution over the whole set of latent variables is determined by the sum of their individual log posterior distributions and then we try to learn a model for which this additive approximation works well.}. This idea is already used in set transformers \cite{settransformer} for combining information from different members of a set. If we modify this suggestion slightly to use an attention-weighted local average, we get a particularly simple form of transformer in which the key, the query and the value are all the same as the embedding itself and the only interaction between locations is attention-weighted smoothing at each level. All of the adaptation occurs in the bottom-up and top-down neural networks at each location, which are depicted by the blue and red arrows in figure \ref{fig:newglomarchitecture}.  These networks are shared across all locations and all time-steps, but possibly not across all levels of the part-whole hierarchy. 

\subsection{Neural fields}
\begin{figure}[h!]
\centerline{\includegraphics[width=4in]{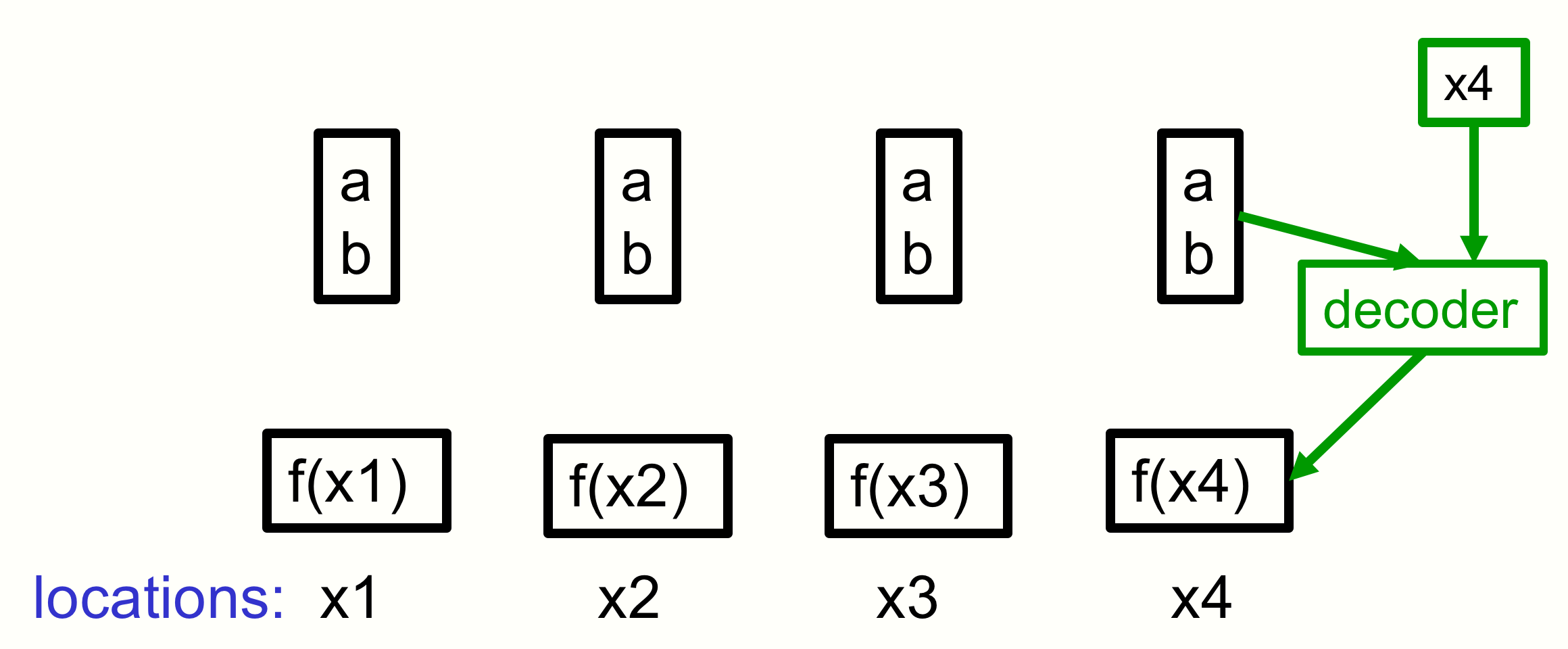}}
\caption{A very simple example of a neural field using individual pixels as the locations. The intensities of four pixels can all be represented by the same code $(a,b)$ even though their intensities vary according to the function $f(x) = ax +b$.  The decoder has an extra input which specifies the location.}
\label{fig:implicit}
\end{figure}

Suppose we want to represent the value of a scalar variable, such as the depth or intensity, at every point in an image. A simple way to do this is to quantize the image locations into pixels and use an array that specifies the scalar variable at each pixel. If the values of different pixels are related, it may be more efficient to use a neural network that takes as input a code vector representing the image and outputs an array of pixel values. This is what the decoder of an autoencoder does. Alternatively we could use a neural network that takes as input a code vector representing the image plus an additional input representing an image location and outputs the predicted value at that location. This is called a neural field \footnote{An early example of using neural fields is described in~\cite{Oore1997}. The "image" is always the same, so only the location input is needed. The 12 channels at each "image" location are the depths returned by 12 sonar detectors pointing in different directions. The match between the neural net's prediction for each location and the actual data at the robot's current location is used to perform a Bayesian update of the mobile robot's probability distribution over locations.} and this way of using neural networks has recently become very popular~\cite{ha2016generating,sitzmann2020implicit,NeRF}. Figure \ref{fig:implicit} shows a very simple example in which the intensities at a set of locations can all be reconstructed from the same code, even though the intensities vary.
 
For a complicated image, the neural net could transform a code vector representing the whole image plus a vector representing an image location into the value at that location. But if images are composed of familiar objects and objects are composed of familiar parts it is much more efficient to use a hierarchy of neural fields\footnote{A small step in this direction is simply to have a separate neural field for each type of object. This makes it easy to represent scenes composed of familiar objects in novel arrangements~\cite{giraffe}.}.  In GLOM, the scene-level top-down neural network converts the scene vector plus an image location into the appropriate object vector for that location. This vector includes information about the 3-D pose of the object relative to the camera. All of the locations that belong to the same object are given exactly the same object-level vector. The object-level top-down neural network then converts an object vector plus a location into the part vector that is appropriate for that location and so on.  For example, exactly the same top-down network acting on exactly the same face vector is able to predict the nose vector in some locations and the mouth vector in other locations. 

\subsection{Explicit versus emergent representations of the part-whole hierarchy}
In the symbolic AI representation of the part-whole hierarchy, each node has a unique symbol or memory address, and this symbol or address has an arbitrary relationship to the content. In GLOM, the embedding vector at a particular level that is shared by all the locations in an island at that level plays the same role as the address of a node in a graph, but this vector is not arbitrary. The top-down neural network must predict the embedding vector of a part at level L from the embedding vector of an object at level L+1. This involves using the pose relative to the camera encoded at level L+1 and a representation of the image location to compute where the location is within the intrinsic coordinate frame of the object. This determines which level L part the location belongs to. 

There is a very important difference between using an address bus to follow a pointer to the representation of a part of a known object and using a top-down neural network to compute the part vector. Unlike table-lookup, the top-down neural net finds it much easier to deal with shapes in which there are symmetries between the parts. Replicated parts, like the legs of a centipede for example, add very little complexity to the neural net and this remains true even if the legs change along the centipede, so long as they change in a predictable way.  Bilateral symmetries that align with the intrinsic coordinate frame of an object reduce the required capacity of the top-down neural net by almost a factor of two.

It is much harder, however, for the neural net to make use of symmetries that do not align with the coordinate frame\footnote{This is why Canonical Capsules~\cite{CanonicalCaps} discover the natural intrinsic coordinate frames.}, and people are generally unaware of such symmetries. Most people, for example, are totally unaware of the threefold rotational symmetry of a cube, despite its name, until they are forced to use a body diagonal through the center of the cube as its intrinsic top-bottom axis~\cite{HintonCube}. They then cease to be aware of any of the right angles in the cube because these angles no longer align with the new intrinsic coordinate frame\footnote{Most people have enormous difficulty imagining a cube when they are forced to use a body diagonal as the top-bottom axis. When asked to point out the corners that are not at the two ends of this axis, they typically point out four corners arranged in a square about halfway up the axis. This structure (two square-based pyramids stuck together base-to-base is actually an octahedron. An octahedron is the dual of a cube with vertices for faces and faces for vertices. So people preserve the fourfold rotational symmetry of a cube relative to its familiar coordinate system. This suggests that the symmetry structure is one of the most important properties encoded in the embedding vector of an object.}. 
 
\section{Some design decisions}
This section discusses some decisions that need to be made when specifying the GLOM architecture.

\subsection{How many levels are there?}
GLOM assumes that the part-whole hierarchy has a fixed depth. People can deal with very deep hierarchies that have stars near the top and atomic nuclei near the bottom. The way to handle such an enormous range is to have a flexible mapping between entities in the world and the levels of GLOM~\cite{HintonPartWhole}. This allows the very same neurons to be used for stars at one time and for atomic nuclei at another, which has the added advantage of facilitating analogies between structures at very different scales like solar systems and atoms. The recursive re-use of the neural hardware raises many interesting issues about temporary storage and control flow~\cite{BaHistory} that will only be fleetingly addressed here.

A reasonable number of embedding levels would be about five. This allows for the pupil and the white of an eye to be the lowest-level sub-sub-parts in a scene composed of objects that are people with parts that are faces and sub-parts that are eyes. If finer details such as the dilation of the iris are required, people probably need to remap the world onto their hardware so that, for example, the face becomes the scene\footnote{The levels in the part-whole hierarchy that are represented in the infero-temporal pathway are probably not the brain's only representation of space. The infero-temporal pathway is used for object recognition and there may well be other representations of the world that are used for other purposes such as detecting ego-motion or visually maintaining balance.}.

One interesting question is whether the bottom-up and top-down neural nets can be shared across levels as well as across locations. This would not work for the lower levels of language processing where entities at different levels, like phonemes or words, have very different properties, but vision is far more fractal.  A big advantage of sharing across levels is that the vector representation used for a particular face when it was at the object level would then be forced to be consistent with its representation when it was at the part level.  This would make it much easier to remap the visual world onto the hardware by simply copying all of the vectors up or down a few levels.  After having used fine details of an eye to extract a highly informed vector representation of the eye when it was at the object level, this same vector could then be used to represent the eye when it was at the sub-part level\footnote{This assumes that the vector can be transported to a different column if the fixation point changes when the face becomes the object of attention rather than the eye.}.

\subsection{How fine-grained are the locations?}
Locations could be as fine-grained as pixels, or they could correspond to larger image patches~\cite{PatchTransformer}. To avoid additional complexity when explaining the basic idea of the paper, I will assume that the grid of locations remains the same at all levels, but this is probably not the best choice. 

The granularity could change at different embedding levels. If higher levels in the part-whole hierarchy use a larger stride,  the top-down neural net would need to output multiple different predictions for the multiple lower level locations that fall within one higher level location. Similarly, the bottom-up neural net would need to look at all the lower-level locations that get combined at the next level up.

One convenient way to be sensitive to a large spatial context whilst also being able to see fine detail is to have images at several different spatial resolutions all of which have the same number of pixels. The coarsest image conveys a large spatial context but lacks fine detail and the finest image conveys the fine details, but only for a small region.  If the visual input is structured into multiple images in this way, it would make sense to make peripheral locations cover larger regions, but this paper will ignore that issue because it makes everything more complicated. 

\subsection{Does the bottom-up net look at nearby locations?}
Even if the granularity at different levels remains unchanged, the bottom-up neural net could look at the embedding vectors at nearby locations.  This is a less pure version of GLOM which allows the interactions between locations to be more complex than just averaging. The purely bottom-up pathway then resembles a convolutional neural net but with the predictions for the next level up being made by a multi-layer neural net that implements a far more complicated function than just a matrix multiply followed by a scalar non-linearity. 

The disadvantage of allowing the bottom-up net to look at other locations is that two locations with identical representations at the part level may have different spatial contexts. We would then lose a very nice property of the pure version of GLOM: locations that have identical representations at the part level make exactly the same bottom-up predictions at the object level. 

By looking at other locations, the bottom-up net can reduce the uncertainty before it predicts a distribution at the next level up and this seems like a good thing to do. But it should be possible to get a similar reduction in uncertainty {\it after} making the prediction when the attention-weighted smoothing combines an uncertain bottom-up prediction from one location with the uncertain bottom-up predictions from nearby locations. Of course, this assumes that the bottom-up net can represent the uncertainty in its predictions and that the uncertainties in different locations can be combined correctly by the attention-weighted smoothing. This issue is addressed in section \ref{sec:uncertainty}.

\subsection{How does the attention work?}
One of the contributors to the update of the embedding of level $L$ at location $x$ is the attention-weighted average of the embeddings of level $L$ at nearby locations at the previous time step. GLOM assumes the simplest form of attention weighting in which the weight $w_{xy}$ that location $x$ gives to the embedding at location $y$ is given by 
\begin{equation}
w_{xy} = \frac{e^{\beta L_x.L_y}}{\sum_z e^{\beta L_x.L_z}}
\end{equation}
where $.$ is the scalar product of the two embedding vectors, $z$ indexes all the locations that location $x$ attends to at level $L$ and $\beta$ is an "inverse temperature" parameter that determines the sharpness of the attention. $\beta$ could increase as GLOM settles to a firm interpretation of the image.  The way attention is intended to work in GLOM has already been used successfully in a system called "ACNe"~\cite{ACNe}.

Pioneering work on using Markov Random Fields for image segmentation \cite{GemanGeman} used the presence of a boundary between pixel $x$ and pixel $y$ to prevent $x$ from attending to $y$. A boundary is more than just a big intensity difference between $x$ and $y$ because its existence depends on the intensities at other locations.
Similarly, early work on learning spatially coherent properties of images used the presence of boundaries to select which expert interpolator to use~\cite{beckerMOE}. Like the seashore, boundaries have a rich life of their own and much more work needs to be done to integrate them into GLOM, especially into its attention mechanism. 

\subsection{The visual input}
In most neural networks, the visual input arrives at the bottom layer. In GLOM, a patch of the raw visual input could define the bottom-level embedding at a location by vectorizing the intensities in the image patch, but it is probably more sensible to first apply a convolutional neural net that can see a larger region of the image. The output of this convolutional net would then be the primary, lowest level embedding at each location. 

The convolutional net is an open loop way to solve the following inference problem: What lowest-level embedding for that location would be able to reconstruct the pixel intensities using the learned neural field shared by all locations. Once the lowest-level embedding has been initialized, it can be refined in a closed loop by backpropagating the reconstruction error through the neural field~\cite{ckiwold}.

There is no need to confine the direct visual input to the primary embedding layer. A coarser scale convolutional net operating on a lower resolution image could provide useful hints about the higher-level embeddings. For example, a pale vertical oval with a darker horizontal band slightly more than halfway up suggests one kind of face~\cite{ViolaJones} so a convolutional net operating on coarse pixels can provide useful information to directly initialize the higher-level embeddings\footnote{The visual part of the thalamus has direct connections to multiple different levels in the hierarchy of visual areas.}.  

\section{Color and texture}
Consider an object whose individual parts are either entirely pale green or entirely mauve. The color of a part is straightforward, but what color is the whole object? One of the motivations for GLOM was the idea that the whole object has a compound color which might be called "pale-green-or-mauve" and {\it at the object level} every location belonging to the object has exactly the same, compound color. The object is pale-green-and-mauve all over. When deciding which other locations at the object level to attend to, preference would be given to locations with a similar compound color.

A similar idea applies to textures.  The individual texture elements have their own shapes and poses and spatial relationships, but an object with a textured surface has exactly the same texture everywhere {\it at the object level}. GLOM extends this idea to shapes. An object may have parts that are very different from one another, but at the object level it has exactly the same compound shape in all of the locations that it occupies. 

\section{Cluster discovery versus cluster formation} 
The EM capsule model~\cite{capsules2018} attempts to activate capsules that represent wholes (e.g. a face) by looking for clusters of similar vector votes for the pose of the whole. These vector votes come from already identified parts (e.g. a nose or mouth) and although the weights on these votes can be modified by an iterative routing procedure the vector votes themselves remain fixed.  This is quite problematic if one of the parts has an under-determined pose.  For example, a circle representing an eye has no specific orientation and its position in a face depends on whether it is a left or right eye. It does, however, provide some information about the scale of the face and it makes a unimodal prediction for the location of the face in the direction orthogonal to the unknown line between the two eyes\footnote{ The Stacked Capsule Autoencoder model~\cite{capsules2019} deals with this issue by using a set transformer~\cite{settransformer} to allow the parts to interact. This should allow the poses and identities of the parts to be disambiguated before they attempt to activate capsules at the next level up.}. 

In GLOM, the embedding vector of a location at level L-1 does not cast an immutable vector vote for the embedding at level L. Instead, it provides a bottom-up vector contribution to this embedding that is combined with the vector contribution coming from level L+1 and the attention-weighted contributions coming from the level L embeddings of other locations to determine the updated level L embedding vector.  The bottom-up contribution can start off being quite vague and it can become sharper from time-step to time-step as top-down and lateral contextual information progressively refines the level L-1 embedding of the location.  The islands of similar embedding vectors at a level can be viewed as clusters, but these clusters are not {\it discovered} in immutable data. They are {\it formed} by the interaction between an intra-level process that favors islands of similarity and dynamically changing suggestions coming from the location's embeddings at adjacent levels.

\section{Replicating embedding vectors over locations}
At first sight, it seems very inefficient to give a copy of the object-level embedding vector to every location that belongs to an object. Compelling intuitions that stem from programming computers with random access memory suggest that it would be much better to have a single copy of a data-structure for the object. These intuitions are probably misleading for neural nets that do not have RAM, and even if RAM is available there are two good reasons for replicating the embedding vectors over an island.

The island growing process at each level may eventually settle down to several islands of near identical vectors, but the search for these islands needs to be able to consider alternative clusterings of locations into islands and it also needs to allow for negotiations between locations within an island about what identical vector to settle on at each level.  These negotiations are non-trivial because each location is also trying to satisfy inter-level constraints that come from its own embedding vectors at the level above and the level below and these embeddings are also being refined at every time-step. During the search, it is very helpful for every location to have its own version of the embedding vector at each level. Uncertainty in the clustering can be represented by making the embedding vector at a location be a blend of the vectors for the different clusters that it might decide to join.  This blend can be refined over time and the fact that it lives in a high-dimensional continuous space should make optimization easier.

Intuitively, a blend of two rather different embedding vectors is not similar to either vector.  This is true in a low-dimensional vector space, but intuitions derived from low-dimensional spaces cannot be trusted when dealing with high-dimensional spaces.  The average of two high-dimensional vectors is much closer to each of those vectors than it is to a random vector. This can be understood by thinking about the correlation between the components of a vector and the components of its average with some other random vector. If the vectors are high-dimensional, this correlation will be very significant~\footnote{ This explains why the first stage of a language model can convert a word like ``bank'' into a single high-dimensional embedding vector rather than needing separate vectors for the ``river'' and the ``money'' senses.}.

A further advantage of islands of near identity is that it allows long range interactions within a level to be sparse. If there is more sparsity at higher levels, the interactions can be longer range without increasing the amount of computation. For locations that belong to an island far away, all the object-level information about that island is contained in each of its locations, so it is only necessary to sample one of those locations for that distant island to compete with other closer islands for a location's attention. Of course, this means that distant islands contribute fewer logits to the attention softmax than closer islands, but the exponential used in the attentional softmax means that one logit from a highly relevant distant island can out-compete multiple logits from a closer but much less relevant island. 

A simple way to choose which other locations are allowed to compete for the attention of location {\bf x} is to sample, without replacement, from a Gaussian centered at {\bf x}. Higher level embeddings can sample the same number of other locations but from a larger Gaussian. The sampling could be done only once so it was part of the architecture of the net. Alternatively, lacunae in the sampling could be greatly reduced by sampling independently at each time step. 

\section{Learning Islands}
Let us assume that GLOM is trained to reconstruct at its output the uncorrupted version of an image from which some regions have been removed. This objective should ensure that information about the input is preserved during the forward pass and if the regions are sufficiently large, it should also ensure that identifying familiar objects will be helpful for filling in the missing regions. To encourage islands of near identity, we need to add a regularizer and experience shows that a regularizer that simply encourages similarity between the embeddings of nearby locations can cause the representations to collapse: All the embedding vectors may become very small so that they are all very similar and the reconstruction will then use very large weights to deal with the very small scale. To prevent collapse, contrastive learning~\cite{BeckerHinton92,Paccanaro2001,Oord2018} uses negative examples and tries to make representations that should agree be close while maintaining separation between representations which should not agree\footnote{Maintaining separation is quite different from asking representations that should be separate to be far apart. Once two representations are sufficiently different there is no further pressure to push them even further apart.}.

Contrastive learning has been applied very successfully to learn representations of image crops~\cite{ting1, Bachman,moco,ting2,tejankar2020isd}  It learns to make the representations of two different crops of the same image agree and the representations of two crops from different images disagree. But this is not a sensible thing to do if our aim is to recognize objects.  If crop 1 contains objects A and B and crop 2 from the same image contains objects B and C, it does not make sense to demand that the representations of the two crops be the same {\it at the object level}. It does make sense at the scene level, however. For scenes containing one prominent object, it may be possible to recognize objects using representations that are designed to recognize scenes, but as soon as we distinguish different levels of embedding it becomes clear that it would be better to use a contrastive loss function that encourages very similar representations for two locations at level L only if they belong to the same entity at level L. If they belong to different level L entities their level L embeddings should be significantly different. 

From the point of view of a location, at all but the top level it needs to decide which other locations its level L embedding should be similar to.  It can then learn to resemble those embeddings and be repelled from the embeddings of locations that belong to different objects in the same or other images. Recent work that uses the similarity of patches along a possible object trajectory to influence whether contrastive learning should try to make them  more similar has shown very impressive performance at finding correspondences between patches in video sequences~\cite{JabriOwensEfros}. 

The obvious solution is to regularize the bottom-up and top-down neural networks by encouraging each of them to predict the consensus opinion. This is the weighted geometric mean of the predictions coming from the top-down and bottom-up networks, the attention-weighted average of the embeddings at nearby locations at the previous time-step the previous state of the embedding. Training the inter-level predictions to agree with the consensus will clearly make the islands found during feed-forward inference be more coherent.

An important question is whether this type of training will necessarily cause collapse if it is not accompanied by training the inter-level predictions to be different for negative examples that use the consensus opinions for unrelated spatial contexts. Using layer or batch normalization should reduce the tendency to collapse but a more important consideration may be the achievability of the goal. 

When the positive examples in contrastive learning are used to try to extract very similar representations for different patches of the same image, the goal is generally not achievable and the large residual errors will always be trying to make the representations collapse. If, however, an embedding at one location is free to choose which embeddings at other locations it should resemble, the goal can be achieved almost perfectly by learning to form islands of identical vectors and attending almost entirely to other locations that are in the same island. This should greatly reduce the tendency towards collapse and when combined with the deep denoising autoencoder objective function and other recent tricks~\cite{grill2020bootstrap,ChenHe2020} it may eliminate the need for negative examples.

\section{Representing coordinate transformations}
When neural networks are used to represent shape, they generally work much better if they represent the details of a shape relative to its intrinsic coordinate frame rather than relative to a frame based on the camera or the world \cite{mocap,NASA}.

Work on the use of neural fields for generating images has established that there are much better ways to represent the location than using two scalars for its x and y coordinates~\cite{sitzmann2020implicit,NeRF}.  The product of a delta function at the location with both horizontal and vertical sine and cosine waves of various frequencies works well. A similar representation is used in transformers for the position of a word fragment in a sentence. 

The success of highly redundant representations of location suggests that there may also be highly redundant representations of the non-translational degrees of freedom of a coordinate transform that work much better in a neural net than the matrices or quaternions commonly used in computer graphics\footnote{The 
  standard matrix representation uses the scale of the matrix to represent the change in scale caused by the coordinate transform. Using the scale of the weights to represent scale in this analog way is a particularly bad idea for neural nets.}.
Let us suppose that we would like the pose of a part ({\it i.e.} the coordinate transform between the retina and the intrinsic frame of reference of a part) to be represented by a vector that is a subsection of the embedding vector representing the part. A multi-layer neural network whose weights capture the viewpoint-invariant coordinate transform between a part and a whole can then operate on the pose vector of the part to predict the pose vector of the whole. If we simply flatten the 4x4 matrix representation of a pose into a vector, it is easy to hand-design a multi-layer neural net that takes this vector as input and produces as output a vector that corresponds to the flattened result of a matrix-matrix multiply, {\it provided we know what matrix to multiply by, which depends on the identity of the part}. This dependence on the part identity was the reason for allocating a separate capsule to each type of part in earlier capsule models. Unfortunately, the vector space of flattened 4x4 matrices does not make it easy to represent uncertainty about some aspects of the pose and certainty about others. This may require a much higher-dimensional representation of pose. Designing this representation by hand is probably inferior to using end-to-end learning with stochastic gradient descent. Nevertheless, section \ref{sec:uncertainty} discusses one approach to representing uncertainty in a neural net, just to demonstrate that it is not a major problem. 

In a universal capsule the part-identity is represented by an activity vector rather than by the choice of which capsule to activate, so the neural net that implements the appropriate part-whole coordinate transform needs to condition its weights on the part-identity vector~\footnote{In
  stacked capsule autoencoders~\cite{capsules2019} the capsule identity determines the default object-part coordinate transform, but the transform can be modulated by a vector that represents the deformation of the object.}. 
Consequently, the entire part-level vector of a location needs to be provided as input to the bottom-up neural net that computes the part-whole coordinate transform. This makes the computation much more complicated but it greatly simplifies the design of the architecture.  It means that we do not need to designate one part of the embedding vector at a level to represent the pose and the rest to represent other aspects of the entity at that level. All we need to do is to make sure that the neural net that predicts the embedding at one level from the embedding below (or above) has sufficient expressive power to apply a coordinate transform to those components of the embedding vector that represent pose and to make this coordinate transform be contingent on those components of the vector that represent the identity of the part. Since this neural net is going to be learned by stochastic gradient descent, we do not even need to keep components of the embedding vector that represent the pose separate from the components that represent other properties of the entity at that level: individual components can be tuned to combinations of pose, identity, deformation, texture {\it etc}.

Entangling the representations of identity and pose may seem like a bad idea, but how else can a bottom-up prediction from a diagonal line express the opinion that the whole is either a tilted square or an upright diamond? To express this distribution using activities of basis functions, we need basis functions that are tuned to combinations of identity and pose.

Using a small matrix or quaternion to represent pose makes it easy to model the effects of viewpoint changes using {\it linear} operations. At first sight, abandoning these explicit representations of pose seems likely to compromise the ability of a capsule to generalize across viewpoints. This would be true if each capsule only dealt with one type of object, but universal capsules will have seen many different types of object from many different viewpoints and any new type of object will be well approximated by a weighted average of familiar types all of which have learned to model the effects of viewpoint. Moreover, the weights in this average will be the same for all viewpoints. So if a novel object is only seen from a single viewpoint, a universal capsule may well be able to recognize it from radically different viewpoints.

The same argument about generalization can be applied to CNNs, but there is a subtle difference:  GLOM is forced to model the coordinate transforms between parts and wholes correctly in order to be able to make use of the spatial relationship between one part and another by using a simple averaging operation at the level of the whole. It is the viewpoint invariance of these part-whole spatial relationships that makes it possible to generalize to radically new viewpoints. 

\section{Representing uncertainty}
\label{sec:uncertainty}
It is tempting to imagine that the individual components of an embedding vector correspond to meaningful variables such as the six degrees of freedom of the pose of an object relative to the camera or the class of an object. This would make it easy to understand the representation, but there is a good reason for making the relationship between physically meaningful variables and neural activities a little less direct: To combine multiple sources of information correctly it is essential to take the uncertainty of each source into account.

Suppose we want to represent M-dimensional entities in such a way that different sources of information can contribute {\it probability distributions} over the M-dimensional space rather than just point estimates.  We could use a population of $N \gg M$ neurons each of which is tuned to a Gaussian in the $M$-dimensional space~\cite{WilliamsAgakov}. If we take logs, a neuron then corresponds to a parabolic bump in the log probability.  This bump could be very wide in some directions and very narrow in others. It could even be a horizontal ridge that is infinitely wide in some of the directions. We treat the activity of a neuron as a vertical scaling of its parabolic bump and simply add up all the scaled bumps to get a parabolic bump which represents the log of the unnormalized Gaussian distribution represented by the population of N neurons.

Sources of information can now contribute probability distributions which will be multiplied together by simply contributing additively to the activities of the N neurons.  If we want to keep $N$ relatively small, there will be limitations on the probability distributions that can be represented, but, given a budget of $N$ neurons, learning should be able to make good use of them to approximate the predictive distributions that are justified by the data. If, for example, it is possible for a part to predict the horizontal location of a whole without making this prediction be contingent on other aspects of the pose or identity of the whole, it would be helpful to tune a handful of the $N$ neurons to well-spaced values on the dimension representing the horizontal location of the whole in the underlying $M$-dimensional space. The part can then contribute a Gaussian distribution along this horizontal dimension by making appropriate contributions to this handful of neurons. The relative magnitudes of the  contributions determine the mean of the Gaussian and their overall scale determines the inverse variance of the Gaussian.

The assumption that the neurons have Gaussian tuning in the underlying $M$-dimensional space of possible entities was just a simplification to show that neural networks have no problem in representing Gaussian probability distributions and combining them appropriately. A much more flexible way to tune the neurons would be to use a mixture of a Gaussian and a uniform~\cite{ContrastiveDivergence}. The log of this distribution is a {\it localized} bump which will be called a unibump. The sides of a unibump splay out and eventually become horizontal when we are far enough from the mean that the uniform completely dominates the Gaussian. Unlike a parabolic bump which has a quadratically large gradient far from its maximum, a unibump has zero gradient far from its maximum so it makes no contribution to the shape of the unnormalized distribution far from its mean. This allows unibumps to represent multi-modal probability distributions. The sum of one set of nearby unibumps can represent one mode and the sum of another set of unibumps that are close to one another but far from the first set can represent another mode. Using neural activities that correspond to vertical scalings of the unibumps, it is possible to control both the location and the sharpness of each mode separately.

The assumption that individual neurons are tuned to a mixture of a Gaussian and a uniform was just a simplification to show that neural networks can represent multi-modal distributions. The basis functions that neurons actually learn for representing multi-modal log probability distributions in an underlying latent space do not need to be local in that space. 

The need to represent uncertainty prevents the simplest kind of representation in which activity in a single neuron represents one dimension of an M-dimensional entity, but it still allows neurons to have tuning curves in the M-dimensional space. Whether it is possible for someone trying to understand the representations to jointly infer both the underlying latent space and the tuning curves of the neurons in that space is a very interesting open problem. But even when it is hard to figure out what the individual neurons are doing it should still be trivial to find islands of nearly identical vectors, so it should be easy to see how GLOM is parsing an image or how a similar model applied to language is parsing a sentence.

When considering how to represent uncertainty about the pose or identity of a part, it is very important to realize that each location assumes that it is only occupied by at most one part at each level of the hierarchy\footnote{We assume the visual world is opaque. Transparency, like reflections in a window of an indoor scene superimposed on an outdoor scene would need to be handled by switching attention between the two different scenes.}. This means that all the neural activities in the embedding vector at a level refer to the same part: there is no binding problem because the binding is done via the location. So a location can use two different neurons whose tuning curves in the underlying M-dimensional space overlap by a lot without causing any confusion.  If we do not start by allocating different subsets of the neurons to different locations, the broad tuning curves in the M-dimensional underlying space that are needed for representing uncertainty will cause confusion between the properties of different objects. That is why coarse coding, which uses a single population of broadly tuned neurons to model several different entities at the same time~\cite{HintonCoarse} cannot model uncertainty efficiently.

\subsection{Combining different sources of information when updating the embeddings}
The embedding at each level is updated using information from the previous time-step at adjacent levels and also at other locations on the same level. These sources are far from independent, especially when the image is static so that the visual input is identical at multiple time-steps. The higher-level embeddings obviously depend on the earlier lower-level embeddings. Also, the same-level embeddings that contribute to the attention-weighted local average will have been influenced by early states of the very embedding that the attention-weighted average is trying to update. To avoid becoming over-confident it is better to treat the different sources of information as {\it alternative} paths for computing the embedding vector from the visual input. This justifies taking a weighted geometric mean of the distributions\footnote{When taking the geometric mean of some distributions we assume that the product of the distributions is renormalized to have a probability mass of 1.} predicted by the individual sources rather than a simple product of these distributions which would be appropriate if they were independent.  
For interpreting a static image with no temporal context, the weights used for this weighted geometric mean need to change during the iterations that occur after a new fixation. Initially the bottom-up source should be by far the most reliable, but later on, the top-down and lateral sources will improve. Experiments with deep belief nets\cite{HintonNSERC2006} show that gradually increasing the weighting of top-down relative to bottom-up leads to more plausible reconstructions at later times, suggesting that this will be important when GLOM is trained as an end-to-end deep denoising autoencoder.   
 \section{Comparisons with other neural net models}
This section compares GLOM to some of the neural net models that influenced its design. 
\subsection{Comparison with capsule models}
The main advantage of GLOM, compared with capsule models~\footnote{Some capsule models already use universal capsules in which vectors of activity rather than groups of neurons are used to represent the part identity, but they do not replicate these vectors across all locations within the object~\cite{RussNitish}.}, is that it avoids the need to pre-allocate neurons to a discrete set of possible parts at each level. The identity of a part becomes a vector in a continuous space of feature activities. This allows for much more sharing of knowledge between similar parts, like arms and legs, and much more flexibility in the number and type of parts belonging to an object of a particular type. 

A second advantage of GLOM is that it does not require dynamic routing. Instead of routing information from a part capsule to a specific capsule that contains knowledge about the relevant type of whole, every location that the part occupies {\it constructs} its own vector representation of the whole. The constraint that a part at one location only belongs to one whole is a necessary consequence of the the fact that the alternative wholes at that location are alternative activity vectors on the same set of neurons. Uncertainty about which of several wholes is the correct parent of a part can still be captured by using blends of activity vectors.

A third advantage of GLOM is that the cluster formation procedure for forming islands is much better than the clustering procedure used in capsule models.  To make methods like EM work well when the number of clusters is unknown, it is helpful to introduce split and merge operations~\cite{UedaSMEM} but these operations happen automatically during island formation.  Hierarchical Bayesian concerns about finding the correct number of clusters at an embedding level are addressed by starting with one island per location and then reducing the number of distinctly different islands by making embedding vectors agree. This reduction occurs in a continuous space with no need for discrete changes in the number of clusters. 

The main disadvantage of GLOM, compared to most capsule models, is that knowledge about the shape of a specific type of object is not localized to a small group of neurons (possibly replicated across quite large regions). Instead, the bottom-up and top-down neural nets (which may be different for every pair of adjacent levels) have to be replicated at every single location. For computer implementations the replication across locations is a big advantage because it allows a weight to be used many times each time it is retrieved from memory, but for biological neural nets it seems very wasteful of synapses. The point of the analogy with genes is that biology can afford to be wasteful so this objection may not be as serious as it seems. There is, however, a more serious issue for a biological version of GLOM: The ubiquitous universal capsules would need to learn the very same knowledge separately at every different location and this is {\it statistically} very inefficient. Fortunately, section \ref{sec:plausible} shows how locations can share what their bottom-up and top-down models have learned without sharing any of their weights.   

By allocating neurons to locations rather than to types of object or part, GLOM eliminates a major weakness of capsule models, but it preserves most of the good aspects of those models:
\begin{itemize}
    \item {\bf Handling the effects of viewpoint properly:} The weights of the bottom-up and top-down neural networks capture the viewpoint-invariant spatial relationships between parts and wholes and the neural activities capture the viewpoint equivariant information about the pose of an object or part.
    \item {\bf Coincidence filtering:} Objects are recognized by using agreement between high-dimensional predictions from their parts. In GLOM, the idea of using agreement is taken even further because it is also used to {\it represent} objects and parts as islands of identity.
    \item {\bf No dynamic allocation of neurons:} The part-whole hierarchy can be represented without dynamically allocating neurons to nodes in the parse tree.
\end{itemize}

\subsection{Comparison with transformer models}
\begin{figure}[h!]
\centerline{\includegraphics[width=5in]{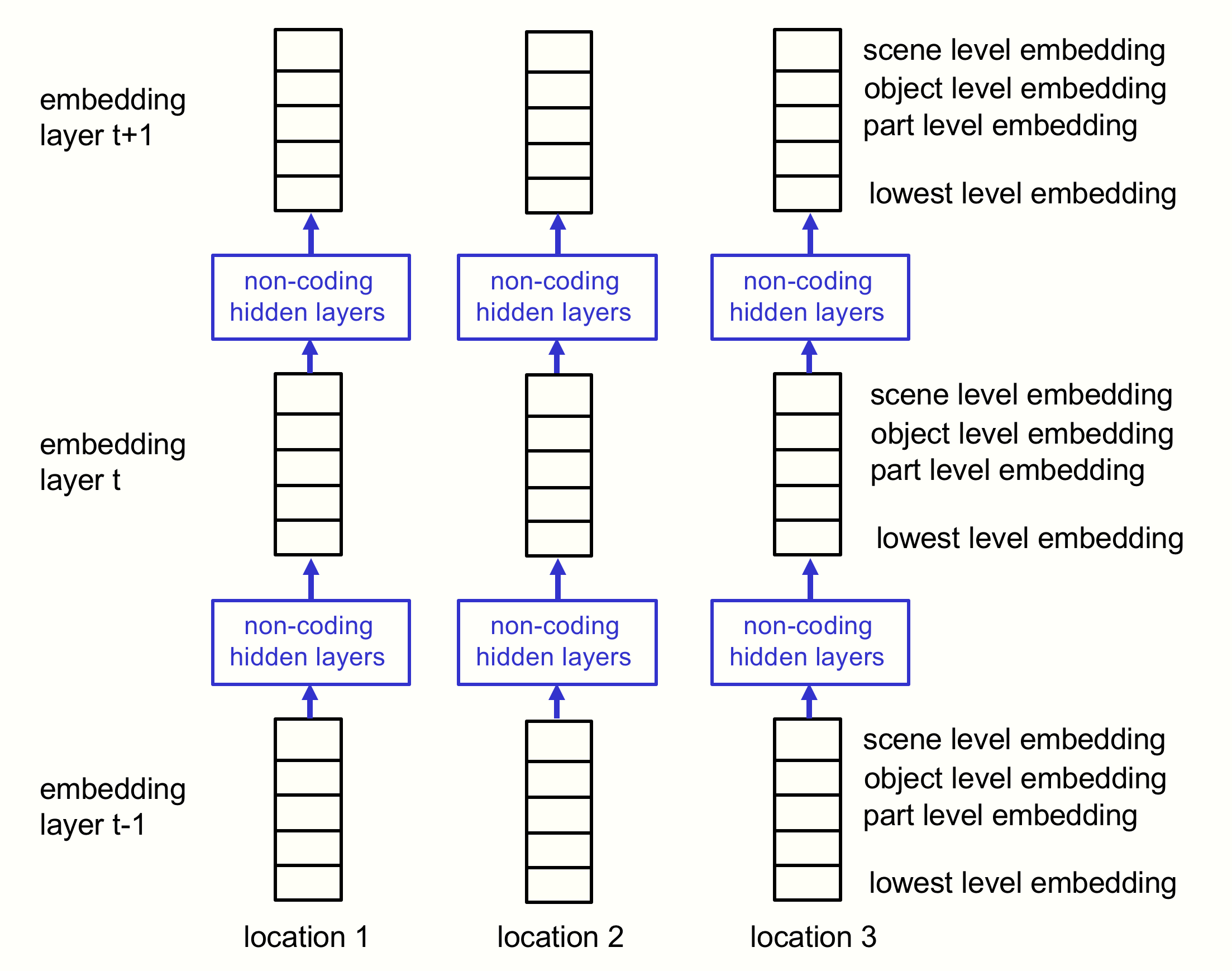}}
\caption{This is a different way of visualizing the architecture shown in figure \ref{fig:newglomarchitecture} which makes the relationship of that architecture to transformers more obvious. The horizontal dimension which represents time in figure \ref{fig:newglomarchitecture} becomes the vertical dimension which represents layers in this figure. At each location, every layer now has embeddings for all of the levels in the part-whole hierarchy. This corresponds to vertically compressing the depiction of the levels within a single time-slice in figure \ref{fig:newglomarchitecture}. A single forward pass through this architecture is all that is required to interpret a static image. All of the level-specific bottom-up and top-down neural nets are shown here as a single neural net. Figure \ref{fig:updown} shows the individual bottom up and top-down neural nets for this alternative way of viewing the GLOM architecture. 
}
\label{fig:oldglomarchitecture}
\end{figure}

The GLOM architecture shown in figure \ref{fig:newglomarchitecture} can be rearranged by viewing the vertical time-slices in figure \ref{fig:newglomarchitecture} as layers in figure \ref{fig:oldglomarchitecture}. This rearrangement of GLOM is then equivalent to a standard version of a transformer~\cite{transformer} but with the following changes:
\begin{itemize}
    \item The weights are the same at every layer because GLOM is a recurrent net and we have converted the time slices into layers.
    \item The attention mechanism is greatly simplified by using the embedding vector at a level as the query, the key and also the value. The complex interactions between different locations that are normally implemented by attention are thus reduced to a simple, attention-weighted, smoothing operation.
    \item The multiple heads used to provide more expressive power in most transformers are re-purposed to implement the multiple levels of a part-whole hierarchy and the interactions between the heads at a location are highly structured so that a level only interacts with the adjacent levels.
    \item The bottom-up and top-down neural networks that compute the interactions between adjacent levels perform coordinate transformations between the distributed representations of the poses of parts and wholes and these coordinate transformations depend on the distributed representations of the types of the part and the whole. 
\end{itemize}
The justification for eliminating the distinction between the query, the key, the value and the embedding itself is as follows: Consider trying to get a potential mouth to be corroborated by a potential nose in a transformer. The mouth needs to ask "is there anyone in the right spatial relationship to me who could be a nose". If so, please tell me to be more mouth-like. This seems to require the mouth to send out a nose query (that includes the appropriate pose relative to the mouth) that will match the key of the nose. The nose then needs to send back a mouth-like value (that includes the appropriate pose relative to the nose). 

But the mouth could also be corroborated by an eye so it needs to send out a different query that will match the key of an eye. This could be handled by using separate heads for a mouth-looking-for-a-nose  and a mouth-looking-for-an-eye (as in categorial grammar), but that seems clumsy.

A more elegant solution (inherited from capsule models) is to use a form of the Hough transform. The potential mouth predicts a vector for the face it might be part of. The potential nose and eye do the same.  All you need now is agreement of the predictions at the face level so query$=$key$=$value$=$embedding. The face level can then give top-down support to its parts instead of the support coming from a value vector sent by one part to another using a coordinate transform specific to the identities of the two parts.
\newpage
 \newpage
\begin{figure}[h!]
\vspace*{0.5in}
\centerline{\includegraphics[width=5in]{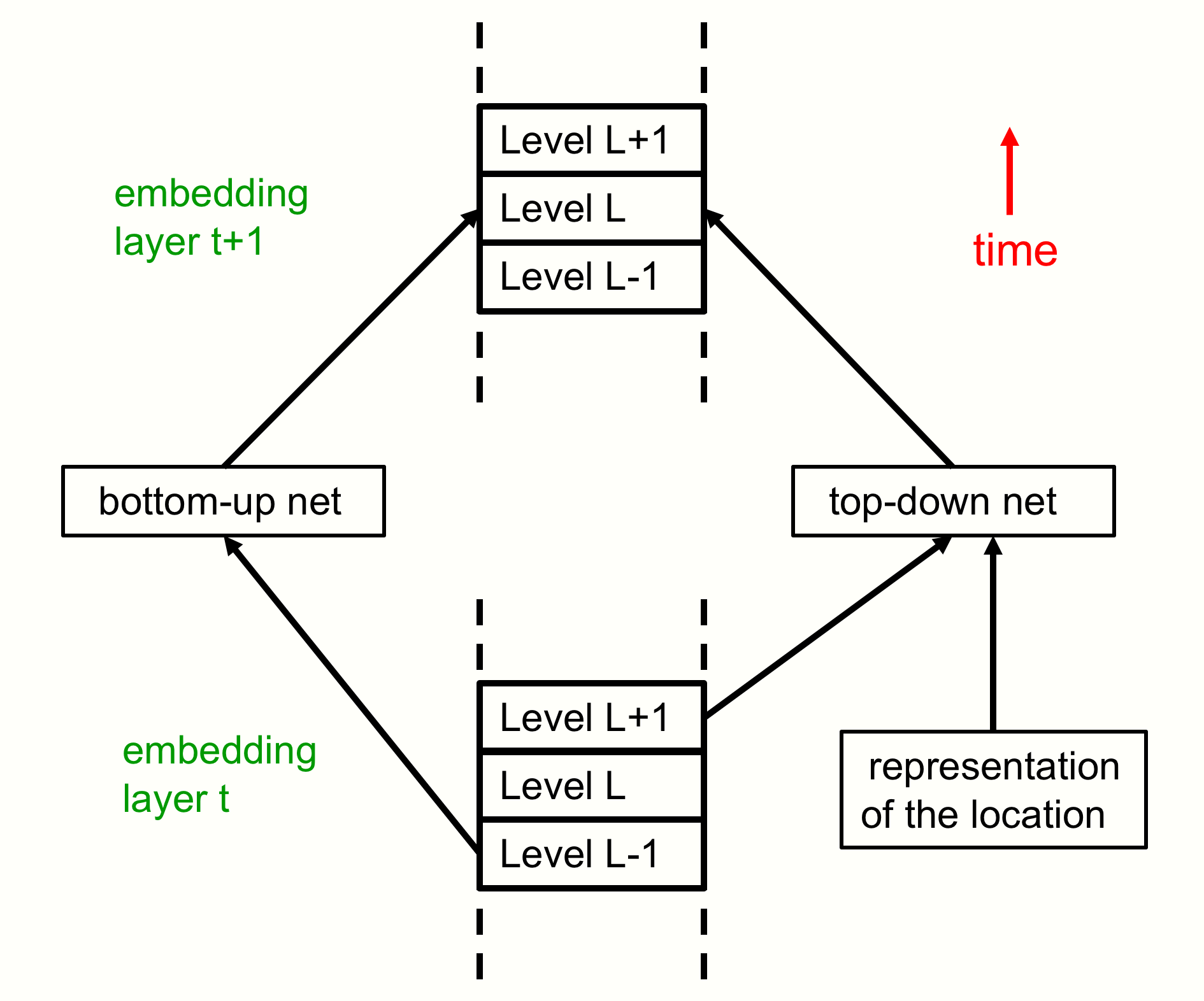}}
\caption{A picture of two adjacent layers of GLOM for a single location ({\it i. e.} part of a single column).  During the forward pass, the embedding vector at level L receives input from the level L-1 embedding vector in the previous layer via a multi-layer bottom-up neural net.  It also receives input from the level L+1 embedding in the previous layer via a multi-layer top-down neural net. The dependence on level L+1 in the previous layer implements top-down effects during the forward pass. The level L embedding in layer $t+1$ also depends on the level L embedding in layer $t$ and an attention-weighted sum of the level L embeddings at other nearby locations in layer $t$. These within-level interactions are not shown.}
\label{fig:updown}
\end{figure}
 \newpage

\subsection{Comparison with convolutional neural networks}
Capsules were originally motivated by three perceived deficiencies of CNNs:
\begin{enumerate}
    \item The pooling operation in a CNN was designed to achieve local {\it invariance} to translation in the activity vector at the next level up. It seems better to ask for invariance in the weights but equivariance in the activities.
    \item CNNs attempt to generalize across viewpoints by using a lot of examples of different viewpoints which may be produced by augmenting the dataset with transformed images. Computer graphics generalizes across viewpoints by having explicit representations of the poses of objects or parts relative to the camera. A change in viewpoint, even a very big one, can be modeled perfectly by {\it linear} operations on these explicit poses.  Using the viewpoint invariant relationship between the pose of a part and the pose of the whole seems like a very efficient way to generalize to radically new viewpoints. CNNs do not appear to be doing this, but appearances can be deceptive.
    \item In CNNs, the activity of a neuron is determined by the scalar product of a weight vector with an activity vector. This is not a good way to model covariance structure which is very important in vision. Taking the scalar product of an activity vector with another activity vector makes powerful operations like coincidence detection and attention much easier. Coincidences in a high-dimensional embedding space are a good way to filter out noise caused by occlusion or missing parts because, unlike sums, they are very robust to the absence of some of the coinciding predictions.
\end{enumerate}

The {\bf first} deficiency is only apparent. It depends on a common misunderstanding of how CNNs represent the positions of parts. The vector of channel activities at a gridpoint can have a rate-coded representation of the position of a part that is far more accurate than the stride between gridpoints. So when the stride is increased at the next level by pooling it does {\it not} mean that the position of a part is encoded less accurately. A bigger stride does not produce representations that are more translationally invariant. Gridpoints are used to allocate neural hardware not to represent positions. Their spacing is limited by the fact that the neural hardware at a gridpoint assumes it will never be representing more than one of whatever it represents, not by the accuracy with which position needs to be represented. 

Attempts to deal with the {\bf second} perceived deficiency led to some interesting models.  The transforming autoencoder~\cite{transformingautoencoder} forced an encoder to extract an explicit representation of pose in each capsule by insisting that the reconstructed image should be a transformed version of the original image and specifying this transformation as a matrix that multiplied whatever matrix elements were extracted by the encoder. Similarly, the EM capsule model extrapolated much better to new viewpoints when it was forced to use a matrix to represent the relationship of a part to a whole.  

Unfortunately, perception has to deal with uncertainties that are not present in computer graphics\footnote{Even if a generative model is stochastic, it may still be certain about which stochastic choices it made. In some more complex generative models, however, a level only specifies the probability distributions of poses for parts at the level below and an iterative process then reconciles these distributions~\cite{OsinderoMRF}. This kind of generative model may be needed for modelling very precise relationships between highly variable parts, such as two praying hands, and it {\it does} need to be able to represent probability distributions over poses.} 
and it needs to be able to represent correlated uncertainties in its pose predictions so that multiple sources of information can be combined properly. This rules out a simple matrix representation of pose. Once we accept that distributions over the possible poses of an entity will be represented by the scales assigned to basis functions in the log probability space, it is quite possible that CNNs actually learn to do something like this. This might allow them to approximate Hough transforms, though this is hard to do without taking scalar products of activity vectors.    

The {\bf third} deficiency can be rectified by moving to a transformer-like architecture that uses scalar products of activity vectors to modulate attention. 

If you like CNNs, GLOM can be viewed as a special type of CNN that differs from a standard CNN in the following ways:
\begin{itemize}
    \item It only uses 1x1 convolutions (except at the front end). 
    \item Interactions between locations are done by parameter-free averaging that implements a coincidence filter which allows it to use a Hough transform to activate units rather than only using matched filters. 
    \item Rather than using a single feed-forward pass through the levels of representation, it iterates to allow top-down influences that are implemented by neural fields.
    \item It includes contrastive self-supervised learning and performs hierarchical segmentation as a part of recognition rather than as a separate task. No more boxes.
\end{itemize}

\subsection{Representing the ISA hierarchy}
An important idea in Good Old-Fashioned Artificial Intelligence (GOFAI) is property inheritance.  It is not necessary to explicitly represent that elephants suckle their young because an elephant ISA mammal and, unless otherwise stated, an elephant inherits this property from its more general type.  A simple way to implement property inheritance in a neural network is to make different entities correspond to different vectors of activity on the same set of neurons. Imagine that the components of the vector that represents a concept are ordered from very general to very specific.  Mammals all have similar values for the more general components and differ on less general components. Indian and African elephants only differ on fairly specific components. When a neural net learns to make the vectors for concepts have causal effects on other vectors, effects which should be the same for all mammals but not the same for all vertebrates will naturally be implemented by the outgoing weights of the neurons that are active for all mammals but not for all vertebrates. This way of implementing property inheritance makes it easy to add exceptions. The components of a vector that are common to birds will learn weights that capture the knowledge that birds fly and the more specific components that differentiate penguins from other birds will learn stronger weights that overrule the general case~\cite{HintonISA}.

This way of implementing property inheritance has the added advantage that types do not need to form a tree.  Dogs inherit many properties from being canines (like wolves)  but they also inherit many properties from being pets (like cats). There is no guarantee that properties inherited from these more general, partially overlapping classes will be consistent, but, unlike logic, neural networks have no difficulty dealing with conflicting evidence. 

At first sight, the idea of using different sections of the vector representation of a concept to capture different levels in the ISA hierarchy conflicts with the idea of using different sections to capture different levels in the part-whole hierarchy.  This seems problematic because {\it hooked beak} is a part of a bird but it also defines a type of bird. The two ideas can be reconciled by first dividing the embedding vector for a location into sections that represent different levels in the part-whole hierarchy and then dividing each section into subsections that represent different levels in the type hierarchy. 

\subsection{The relationship to 2-D Ising models}
For each location separately, the embedding vectors at levels L-1 and L+1 on the previous time-step provide input to the neurons that represent the current embedding vector at level L. This acts like the conditioning input in a conditional Markov Random Field: it influences the current step of the iterative, island forming process that tries to make the embedding of the location at level L agree with the embeddings of other locations at level L. 

In a 2-D Ising model, a two-dimensional array of binary-valued spins settles into a state in which nearby spins tend to agree so as to minimize an energy function that favors agreement between neighboring spins.  The model proposed here resembles the 2-D Ising model because it uses a 2-D  grid of image locations but it generalizes the model in the following ways:
\begin{enumerate}
\item It replaces binary spins with high-dimensional real-valued vectors. The fact that these lie in a continuous space should facilitate the search for islands of agreement.
\item it replaces a single field of spins with fields at multiple levels, and allows adjacent level embeddings of the same location to interact \cite{HeZemelPerpinan,Saremi}. The interactions between levels are quite complicated because they involve coordinate transformations between parts and wholes. So for each pair of adjacent embedding levels, the top-down and bottom-up interactions at each location must be computed by a multi-layer neural net rather than a simple weight matrix.
\end{enumerate}

\subsection{Comparison with other methods for removing redundancy}
Methods like principal components analysis remove redundancy in the data by limiting the number of available dimensions in the representation. By contrast, a restricted Boltzmann machine with a large number of hidden units squeezes out redundancy by making nearly all of the exponentially many possible binary configurations of the hidden units have such high energy that they are effectively unavailable. This is a much more flexible way of eliminating redundancy~\cite{Zhu}. It can model multiple fat manifolds\footnote{A manifold is a subset of the points in a space that have lower intrinsic dimensionality than the full space. If we take the points on a manifold and add a small amount of noise that has full dimensionality, the points no longer form a strict manifold, but they will all be close to the manifold. Such a set of points will be said to lie on a fat manifold.} that have different intrinsic dimensionalities and even within a fat manifold it can model variations in the effective dimensionality in different parts of the manifold. 
The island forming objective belongs to the second class of methods. At each level, it allows for a large number of small islands if that is what the data requires but strives to use a small number of large islands if that is possible. 
 \section{Video}
This paper focuses on using the GLOM architecture to process a single fixation of a static image, but the architecture is motivated by the need to deal with video and learning from video is often much easier than learning from static images~\cite{flowcaps}, so I will briefly discuss the simplest temporal extension which is to a single fixation of a time-varying image. 

To avoid confusion it may be helpful to distinguish three different types of time:
\begin{itemize}
    \item {\bf Event time:} This is the actual time at which an event occurs.
    \item {\bf Representation time:} This is the actual time at which a particular representation of an event occurs in the neural network. If the bottom-up neural network uses a predictive model, representations of events could be in synchrony with the events themselves or they could even precede the events which would make catching a ball a lot easier.
    \item {\bf Reference time:} This is the actual time that an internal representation refers to. When a memory is retrieved, for example, the reference time of the constructed representation is usually long before the representation time. The reference time can also differ by a lot from the event time if the memory is not veridical. 
\end{itemize}

For a sequence of frames representing a static image, multiple time steps can be used to settle on an appropriate set of islands at each level. But in a dynamic image, the very same time-steps must also be used to deal with the fact that the occupants of a location at each level can change with time.

An advantage of using islands of identical vectors to represent an object is that motions between successive frames that are small compared to the size of the object only require large changes to a small subset of the locations at the object level. All of the locations that remain within the object need to change only sightly to represent the slight change in pose of the object relative to the camera. 

If the changes in an image are small and predictable, the time-steps immediately following a change of fixation point can be used to allow the embeddings at all levels to settle on slowly changing islands of agreement that track the changes in the dynamic image. The lowest level embeddings may change quite rapidly but they should receive good top-down predictions from the more stable embeddings at the level above. Once the embeddings have formed sensible islands, there is then no problem in using the very same time-step for improving the interpretation of each frame and for keeping the embeddings locked on to the dynamic image.

If the changes are rapid, there is no time available to iteratively settle on a good set of embedding vectors for interpreting a specific frame. This means that the GLOM architecture cannot correctly interpret complicated shapes if the images are changing rapidly.  Try taking a irregularly shaped potato and throwing it up in the air in such a way that it rotates at one or two cycles per second. Even if you smoothly track the potato, you cannot see what shape it is.

 \section{Is GLOM biologically plausible?}
\label{sec:plausible}
Although GLOM is biologically inspired, it has several features that appear to make it very implausible as a biological model. Three of these features are addressed here.
\begin{itemize}
    \item The weight-sharing between the bottom-up or top-down models in different columns.
    \item The need to process negative pairs of examples for contrastive learning without interrupting the video pipeline.
    \item The use of backpropagation to learn the hidden layers of the top-down and bottom-up models.
\end{itemize}
\subsection{Is the neocortex a giant distillery?}
The replication of DNA in every cell is unproblematic: that is what DNA is good at. But biologists often object to models that use weight-sharing claiming that there is no obvious way to replicate the weights~\cite{naturereviewsBP}. GLOM, however, suggests a fairly simple way to solve this problem by using contextual supervision.  In a real brain, what we want is an efficient way of training the bottom-up and top-down nets at a location so that they compute the same function as the corresponding nets at other locations. There is no need for the weights to be identical as long as corresponding nets are functionally identical. We can achieve this using knowledge distillation~\cite{Caruana2006,distillation}. For each level separately, the two students at each location are the bottom-up and top-down neural nets.  The teacher is the consensus opinion that is a weighted geometric mean of the opinions of the two students, the previous state of the embedding, and the attention-weighted embeddings at {\it other} locations\footnote{Strictly speaking, this is an example of co-distillation where the ensemble of all the students is used as the teacher. Co-distillation was initially based on an analogy with how scientists in a community learn from each other \cite{codistill}, but the same mechanism could be used in a community of columns. In both cases it would help to explain how a system can win by just replicating a lot of people or columns without requiring any significant architectural innovation.}.
 
Regressing a student's prediction towards the consensus, allows knowledge in the neural nets at other locations to be transferred to the student via the attention weighted averaging. It is not as effective as sharing weights with those other neural nets, but it works quite well~\cite{distillation} and in the long run all of the networks will converge to very similar functions if the data distribution is translation invariant. In the long run, however, we are all dead\footnote{There are alternative facts.}. So it is interesting to consider what happens long before convergence when the local models are all fairly different.

Suppose all of the locations that form a nose have the same embedding vector at the part level. If they all had exactly the same bottom-up model, they would all make exactly the same prediction for the face at the object level. But if the bottom-up models at different locations are somewhat different, we will get a strong ensemble effect at the object level: The average of all the simultaneous bottom-up predictions for the same object in different locations will be much better than the individual predictions.

One advantage of sharing knowledge between locations via distillation rather than by copying weights is that the inputs to the bottom-up models at different locations do not need to have the same structure. This makes it easy to have a retina whose receptive fields get progressively larger further from the fovea, which is hard to handle using weight-sharing in a convolutional net. Many other aspects, such as the increase in chromatic aberration further from the fovea are also easily handled. Two corresponding nets at different locations should learn to compute the same function of the optic array even though this array is pre-processed differently by the imaging process before being presented to the two nets. Co-distillation also means that the top-down models do not need to receive their location as an input since it is always the same for any given model.

Finally, using distillation to share knowledge between location specific neural networks solves a puzzle about the discrepancy between the number of synapses in the visual system (about $10^{13}$) compared to the number of fixations we make in our first ten years (about $10^9$). Conservative statisticians, concerned about overfitting, would prefer these numbers to be the other way around\footnote{It should help that each example allows contrastive learning at several different levels and the target vectors for contrastive learning are much richer than a single one-of-N label}. If we use, say, $10^4$  columns in different locations, the bottom-up and top-down models at one location only have about $10^9$ synapses between them. Conversely, the number of training examples used to learn the knowledge that is shared across an ensemble of $10^4$ locations is about $10^{13}$, though many of these examples are very highly correlated.

Neural networks that have more training cases than parameters are less magical than some of the highly over-parameterized networks in current use but they may generalize in more predictable ways when presented with data that lies outside their training distribution because the function they compute has been much more highly constrained by the data. 

\subsection{A role for sleep in contrastive learning?}
If negative examples are required, GLOM might appear less plausible as a biological model because of the added complexity of finding and processing pairs of images that are similar when they should not be. There is, however, one intriguing possibility that emerged from conversations with Terry Sejnowski in 1983 and 2020. 

When using contrastive learning to get representations that are similar for neighbouring video frames, the most effective negative examples are frames in the same video that are nearby but not immediately adjacent. We could avoid compromising the real-time performance of GLOM by taking it offline at night to do the negative learning that prevents the representations from collapsing. If the highest level embeddings have the ability to generate sequences at the highest level, the top-down networks could be used to generate sequences of embeddings at every level in each column. This process does not require any attention between columns because it does not need to perform perceptual inference, so it might be able to generate plausible sequences at a much faster rate than the normal speed. Then we simply do the negative learning for the bottom-up models using the same length of real-time window as is used when awake.  There is evidence that high-speed, top-down sequence generation occurs during the spindle stage of sleep~\cite{LeeWilson,BuzReplay}. 

The idea that sleep is used to keep apart representations that should not be confused is not new~\cite{CrickMitchison}. Hinton and Sejnowski \cite{BMsleep} even suggested that sleep could be used for following the derivatives of the normalizing term in the negative phase of Boltzmann machine learning. But this reincarnation of the idea has two big advantages over Boltzmann machines.  First, contrastive unsupervised learning scales much better than Boltzmann machine learning and second, it is far more tolerant of a temporal separation between the positive and negative phases. 

Preliminary experiments using contrastive learning for MNIST digits show that the learning still works if a large number of positive updates are followed by a large number of negative updates. Representation collapse is fairly slow during the positive-only learning and the representations can shrink by a significant factor without much affecting performance. So maybe some pairs of embeddings that ought to be well separated get too close together during the day and are then pushed apart again at night. This would explain why complete sleep deprivation for a few days causes such serious mental confusion\footnote{If sleep is just for doing extra rehearsal or for integrating the day's experiences with older experiences, it is not clear why complete lack of sleep for a few days has such devastating effects.}. The experiments with MNIST also show that after a lot of positive-only learning, performance degrades but is rapidly restored by a small amount of negative learning.

To avoid very long periods of negative-only learning, it might be advisable to start with a negative phase of sleep to push representations apart, and then alternate with a positive phase using input sequences that were generated from the top level or even from a recurrent network close to the sensory input. This conflicts with the Crick-Mitchison theory that REM sleep is for unlearning,  but it would still be compatible with our failure to remember almost all of our dreams if episodic memory retrieval depends on the top-level, and the top-level simply does not learn during REM sleep because those episodes simply did not happen.  

\subsection{Communicating error derivatives in the brain}
The straightforward way to train GLOM is to ask it to fill in missing regions of images and to backpropagate the reconstruction error through the entire temporal settling process using backpropagation through time. The contrastive representation learning at each level can then be viewed as an additional regularizer.  Unfortunately, it is hard to see how a brain could backpropagate through multiple time steps.  If, however, the consensus opinion at every level can provide a sufficient teaching signal for the bottom-up and top-down models that predict the embedding vector at that level, implementation in a brain becomes a lot more feasible. 

If we could ensure that the representations improved over time, the temporal derivatives of neural activities could represent error derivatives and the local learning procedure would then be spike-time dependent plasticity in which the increase in a synapse strength is proportional to the product of the pre-synaptic activity with the post-synaptic rate of increase of activity.\footnote{This fits very well with the strongly held beliefs of Jeff Hawkins and others that the brain learns by predicting what comes next.} Assuming spikes are caused by an underlying rate variable, we can get a noisy but unbiased estimate\footnote{Stochastic gradient descent is extremely tolerant of noise in the gradient estimates so long as they are unbiased.} of the rate of change of this underlying rate variable by applying a derivative filter to the post-synaptic spike train, which is exactly what STDP does. 

A recent review paper~\cite{naturereviewsBP} discusses at great length how temporal derivatives can be used as error derivatives in order to approximate backpropagation in a feedforward network\footnote{This paper summarizes the results of simulations that show that this proposal can be made to work quite well, but not as well as vanilla CNNs. A significant contributor to the performance gap is the statistical inefficiency caused by the lack of weight-sharing and co-distillation should help to fix this.}. The review paper assumes a separate phase in which derivatives, in the form of activity perturbations, are allowed to flow back from the higher levels  to the lower levels. This process does not seem plausible for a video pipeline.  By contrast, the settling process of GLOM propagates the derivatives required for learning as the temporal derivatives of activity at all levels and the time steps required for this propagation can be the very same time steps as are used for video frames. 

For dynamic images, it may seem paradoxical that the representations just keep getting better, but it is no more paradoxical than a surfer who just keeps going downhill without ever changing her elevation. The surface on which the surfer is going downhill is not the same surface. Similarly the time-slice of reality for which the representations are forever improving is not the same time slice. The brain surfs reality.

Unfortunately, this does not explain how to get the derivatives required for learning the hidden layers of the bottom-up and top-down neural networks. Nor does it explain how the derivatives of the error signals at each level are backpropagated through the bottom-up or top-down networks to make the appropriate contributions to the derivatives at adjacent levels. Those thorny issues are addressed in another paper which is in preparation.
 \section{Discussion}
This paper started life as a design document for an implementation but it was quickly hijacked by the need to justify the design decisions. I have used the imaginary GLOM architecture as a vehicle for conveying a set of interconnected ideas about how a neural network vision system might be organized. The absence of a working implementation makes it easier to focus on expressing the ideas clearly and it avoids the problem of confounding the quality of the ideas with the quality of the implementation, but it also creates serious credibility concerns. The difference between science and philosophy is that experiments can show that extremely plausible ideas are just wrong and extremely implausible ones, like learning a entire complicated system by end-to-end gradient decent, are just right. I am currently collaborating on a project to test out the ability of the GLOM architecture to generalize shape recognition to radically new viewpoints and I am hoping that other groups will also test out the ideas presented here. This paper has gone on long enough already so I will conclude by making some brief philosophical comments.

The idea that nodes in a parse tree are represented by islands of similar vectors unifies two very different approaches to understanding perception. The first approach is classical Gestalt psychology which tried to model perception by appealing to fields and was obsessed by the idea that the whole is different from the sum of the parts\footnote{Thanks to George Mandler for this more accurate translation.}. In GLOM, a percept really is a field and the shared embedding vector that represents a whole really is very different from the shared embedding vectors that represent the parts. The second approach is classical Artificial Intelligence which models perception by appealing to structural descriptions. GLOM really does have structural descriptions, and each node in the parse tree has its own "address" but the addresses live in the continuous space of possible embeddings, not in the discrete space of hardware locations.

Some critics of deep learning argue that neural nets cannot deal with compositional hierarchies and that there needs to be a "neurosymbolic" interface which allows neural network front- and back-ends to hand over the higher-level reasoning to a more symbolic system\footnote{This is reminiscent of Cartesian dualism which postulated an interface between the body and the mind.}. I believe that our primary mode of reasoning is by using analogies which are made possible by the similarities between learned high-dimensional vectors, and a good analogy for the neurosymbolic interface is a car manufacturer who spends fifty years expounding the deficiencies of electric motors but is eventually willing to use them to inject the gasoline into the engine. 

The phenomenal success of BERT~\cite{BERT}, combined with earlier work that demonstrates that neural networks can output parse trees if that is what the task requires~\cite{grammar}, clearly demonstrates that neural networks can parse sentences if they want to.  By structuring the interactions between the multiple heads in BERT so that they correspond to levels of representation and by adding a contrastively learned regularizer to encourage local islands of agreement over multiple word fragments at each level, it may be possible to show that GLOMBERT actually does parse sentences.

 \subsubsection*{Acknowledgments}
Many people helped me arrive at the set of ideas described in this paper.  Terry Sejnowski, Ilya Sutskever, Andrea Tagliasacchi, Jay McClelland, Chris Williams, Rich Zemel, Sue Becker, Ruslan Salakhutdinov, Nitish Srivastava, Tijmen Tieleman, Taco Cohen, Vincent Sitzmann, Adam Kosoriek, Sara Sabour, Simon Kornblith, Ting Chen, Boyang Deng and Lala Li were particularly helpful. People who helped me to improve the presentation of the ideas include David Fleet, David Ha,  Michael Isard, Keith Oatley, Simon Kornblith, Lawrence Saul, Tim Shallice, Jon Shlens, Andrea Tagliasacchi, Ashish Vaswani and several others. I would especially like to thank Jeff Dean and David Fleet for creating the environment at Google that made this research possible. There are probably many highly relevant papers that I should have read but didn't and I look forward to learning about them. 
\bibliographystyle{apalike}
\bibliography{glom}
\end{document}